# Discretizing SO(2)-Equivariant Features for Robotic Kitting

Jiadong Zhou[1], Yadan Zeng[1], Huixu Dong[2], and I-Ming Chen[1]

*Abstract*—Robotic kitting has attracted considerable attention in logistics and industrial settings. However, existing kitting methods encounter challenges such as low precision and poor efficiency, limiting their widespread applications. To address these issues, we present a novel kitting framework that improves both the precision and computational efficiency of complex kitting tasks. Firstly, our approach introduces a fine-grained orientation estimation technique in the picking module, significantly enhancing orientation precision while effectively decoupling computational load from orientation granularity. This approach combines an SO(2)-equivariant network with a group discretization operation to preciously predict discrete orientation distributions. Secondly, we develop the Hand-tool Kitting Dataset (HKD) to evaluate the performance of different solutions in handling orientation-sensitive kitting tasks. This dataset comprises a diverse collection of hand tools and synthetically created kits, which reflects the complexities encountered in real-world kitting scenarios. Finally, a series of experiments are conducted to evaluate the performance of the proposed method. The results demonstrate that our approach offers remarkable precision and enhanced computational efficiency in robotic kitting tasks.

## I. INTRODUCTION

The precision of pick-and-place operations is truly crucial for effective robotic handling in logistics and industrial settings, especially for kitting tasks [1]. Kitting refers to the process of selecting various items and accurately positioning them in designated spots within a kit. The ability to manipulate objects with high precision has a direct influence on the reliability, efficiency and overall success of automated kitting systems. However, a critical challenge emerges when robots are required to kit objects with precise position and orientation, especially when dealing with items of irregular shapes, large aspect ratios, or kits designed with closely fitting cavities (Fig. 1). This complexity underscores the necessity for advanced solutions to improve the precision of robotic kitting operations.

Robotic kitting has gained significant research interest, typically approached as a pick-and-place problem. Traditional methodologies [2]-[6] rely on accurate pose estimation of objects, followed by additional planning of pick and place operations. These methods require extensive prior knowledge of objects (e.g. 3D models and human annotations) and considerable manual tuning for object-specific planning. In contrast, recent works [7]-[10] have shifted toward action-centric methodologies that streamline

*The project is supported by A*STAR under "RIE2025 IAF-PP Advanced ROS2-native Platform Technologies for Cross sectorial Robotics Adoption (M21K1a0104)" programme.

The authors are with [1]School of Mechanical and Aerospace Engineering, Nanyang Technological University, Singapore, and [2]Grasp Lab, Zhejiang University, China. {jiadong001, yadan001}@e.ntu.edu.sg, huixudong@zju.edu.cn, michen@ntu.edu.sg

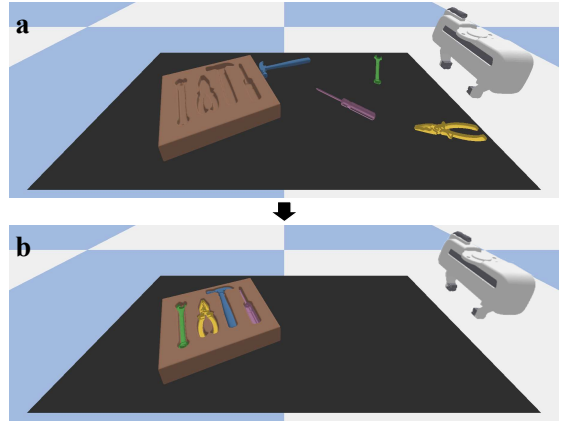

Fig. 1. Simulation of a hand-tool kitting task with a parallel-jaw gripper. The goal is to (a) pick up tools randomly positioned on a table and (b) place them into their corresponding cavities in the kit block.

system architectures and generalize across various manipulation tasks, including kitting. These strategies utilize fully convolutional neural networks (FCNNs) for an end-to-end mapping from image pixels to actions. A prime example is the Transporter Network [8], which produced dense probability maps for object placements via cross-correlation between encoded image crops and scene encodings. Subsequent advancements [11], [12] have refined these methods by integrating steerable convolutional kernels [13]-[16], extending the translational-equivariance of FCNNs to include rotations, achieving SE(2)-equivariance. Huang et al. [17] proposed the Equivariant Transporter, built upon [8] by incorporating bi-equivariance associated with discrete rotation groups, and theoretically advanced to continuous rotation groups [18]. However, despite achieving pixel-level positional precision, these methodologies often exhibit limited orientation precision, typically setting orientation numbers to low values such as 20 [10], 32 [11] or 36 [8], [17]. Within these action-centric frameworks, attempts to increase orientation granularity by upscaling orientation numbers lead to significantly larger model sizes. Furthermore, some researchers [19]-[21] leverage coarse-to-fine strategies for enhanced precision. Valassakis et al. [21] combined pose estimation with end-to-end learning for sub-millimeter precision, and Sóti et al. [19] pursued iterative refinement to achieve precise orientation control. Although these methods increase manipulation precision, they result in a decrease in operational speed. In addition, there is a noticeable lack of benchmarks that address the high orientation sensitivity in complex kitting tasks. Current kitting datasets often target subproblems, such as object recognition [22] or estimating only placement poses [6], [20]. Even as general manipulation benchmarks such as Raven-10 [8] and RLBench [23] offer tasks of varying complexities, they fall short in adequately tackling orientation sensitivity.

Unfortunately, existing research in robotic kitting reveals a significant gap in addressing tasks with high orientation sensitivity, both in methodologies and benchmarks. Traditional approaches based on pose estimation and coarse-to-fine strategies are impeded by their intricate system designs and limited computational efficiency. Meanwhile, action-centric approaches struggle to achieve high orientation precision without compromising system efficiency. The critical challenge, therefore, is to precisely handle object orientations while simultaneously ensuring overall system efficiency. In response, our research aims to refine the precision of kitting tasks, with a particular focus on achieving enhanced orientation precision in conjunction with computational efficiency.

Motivated by the unresolved issues in the aforementioned areas, in this work, we present a novel kitting framework inspired by the Equivariant Transporter [17], which improves both precision and computational efficiency for orientation-sensitive kitting tasks. Firstly, we distinctively introduce a fine-grained orientation estimation strategy into the picking module. This strategy integrates an SO(2)-equivariant network with a group discretization operation. The equivariant design of the network not only boosts sample efficiency via weight sharing but also enables the representation of continuous orientation characteristics as a band-limited Fourier series. A group discretization operation is applied subsequently, which samples this continuous orientation representation at designated angles to produce discrete orientation distributions. By adjusting the sampling rate, we achieve flexible control over orientation precision without compromising the network's structural integrity.

Secondly, we introduce the Hand-tool Kitting Dataset (HKD), meticulously designed to reflect the orientation-sensitive challenges of real-world hand-tool kitting tasks. It comprises a diverse set of hand tools, each with unique geometries and large aspect ratios, as well as synthetically generated kits with tightly conformal cavities. Coupled with a simulation platform that replicates the complete kitting processes, as depicted in Fig. 1, the HKD facilitates a fair and straightforward comparison among different approaches. By creating this dataset, we aim to set a new benchmark in the field, promoting progress in industrial kitting applications.

Finally, extensive experiments on the HKD have demonstrated the remarkable precision and computational efficiency of our methodology. These results highlight the effectiveness of our methods in addressing orientation-sensitive kitting tasks and validate the HKD as a reliable benchmark for evaluating such challenges. Additionally, subsequent experiments on modified Raven-10 tasks showcase our approach's ability to generalize across tasks with varying levels of orientation sensitivity. Consequently, our approach enhances orientation precision in an end-to-end fashion while concurrently optimizing computational efficiency, marking a significant advancement in the field of robotic kitting.

In summary, our main contributions include:
- A kitting framework that incorporates a fine-grained orientation estimation method for enhanced precision and efficiency in the picking module.
- A Hand-tool Kitting Dataset that benchmarks orientation-sensitive kitting solutions and accurately reflects the challenges of real-world kitting scenarios.
- Extensive experiments that validate the outstanding performance of the proposed approach.

## II. BACKGROUNDS

### A. Planar Rotation Groups

We focus on planar rotations characterized by SO(2) and its cyclic subgroup $C_N$. SO(2) is a continuous group of all rotations, defined as $\{r_\theta \mid \theta \in [0, 2\pi)\}$. The discrete group $C_N = \{r_{\theta_i} \mid \theta_i \in 2\pi i/N, i \in [0, N)\}$ consists of N distinct rotations by multiples of $2\pi/N$.

### B. Group Representation

A linear representation $\rho : G \to GL(\mathbb{R}^d)$ maps each group element $g \in G$ to an invertible $d \times d$ matrix $\rho(g)$, with a representation dimension $d_\rho = d$. We consider four representations of SO(2) and $C_N$ that describe how vectors transform under rotation. The *trivial representation* $\rho_0$, defined by $\rho_0(g) = 1$ for all $g \in G$, ensures any vector remains invariant by rotation. The *standard representation* $\rho_1$ maps each element to its standard $2 \times 2$ rotation matrix ($j = 1$ in Equation 1), indicating conventional rotation in $\mathbb{R}^2$. The *regular representation* $\rho_{reg}$ of $C_N$ acts on vectors in $\mathbb{R}^N$ by cyclically permuting vector elements, denoted as $\rho_{reg}(r_{\theta_i})(v_0, v_1, \cdots, v_{N-1}) = (v_{N-i}, \cdots, v_{N-1}, v_0, v_1, \cdots, v_{N-i-1})$, where $r_{\theta_i}$ is the *i*-th element in $C_N$. The *irreducible representation* (irrep) serves as the foundational element for constructing any representation in the group. For SO(2), the real-valued irreps $\rho_j$ include the trivial representation $\rho_0$ and $2 \times 2$ rotation matrices with frequencies $j \in \mathbb{N}^+$, denoted as:

$$\rho_j(r_\theta) = \begin{bmatrix} \cos(j\theta) & -\sin(j\theta) \\ \sin(j\theta) & \cos(j\theta) \end{bmatrix}. \quad (1)$$

The irreps of $C_N$ are identical to those of SO(2) up to a frequency of $(N-1)/2$.

### C. Feature Vector Fields

Images and feature maps are interpreted as feature vector fields, denoted by $f : \mathbb{R}^2 \to \mathbb{R}^{d_\rho}$, where each spatial position $x \in \mathbb{R}^2$ is mapped to a feature vector $f(x) \in \mathbb{R}^{d_\rho}$. An element $g \in SO(2)$ acts on the field *f* via a representation $\rho$ by rotating pixel positions in the spatial domain $\mathbb{R}^2$, coupled with a transformation in the channel domain $\mathbb{R}^{d_\rho}$ according to $\rho$. This group action is formally expressed as:

$$\left[T_g^\rho(f)\right](x) = \rho(g) \cdot f(\rho_1(g)^{-1} x). \quad (2)$$

Note that feature fields are inherently associated with group representations: a "regular field" is associated with a regular representation, while an "irrep field" is linked to an irrep.

## D. Equivariant Mapping

A function $F$ is considered equivariant with respect to a group action if it satisfies the following condition:

$$F(T_g^{\rho_{\text{in}}}[f]) = T_g^{\rho_{\text{out}}}[F(f)]. \quad (3)$$

This implies that transforming the input $f$ using $T_g^{\rho_{\text{in}}}$ and then applying the function $F$ yields the same result as first applying $F$ to $f$, followed by the transformation of the resulting output by $T_g^{\rho_{\text{out}}}$. When $\rho_{\text{out}} = \rho_0$, adopting the trivial representation, the function $F$ is deemed rotation invariant, illustrating a special case of rotation equivariance.

## III. METHODOLOGY

### A. Problem Statement

This study focuses on the effective learning of kitting operations requiring high precision. We address kitting challenges within a planar pick-and-place framework, utilizing a parallel-jaw gripper to pick objects and accurately place them in designated spots, ensuring precision in both position and orientation. The task is formulated as follows: given a visual observation $o_t$ of the workspace, the objective is to predict two probability distributions: $p(a_{\text{pick}} | o_t)$ for pick actions, and $p(a_{\text{place}} | o_t, a_{\text{pick}})$ for place actions conditioned on $a_{\text{pick}}$. The actions $a_{\text{pick}} \in \text{SE}(2)$ and $a_{\text{place}} \in \text{SE}(2)$ represent the gripper poses for executing pick and place actions respectively. The actions that maximize the respective probabilities, denoted as $a_{\text{pick}}^*$ and $a_{\text{place}}^*$, are chosen for performing the kitting operation. The visual observation $o_t$ is an orthographic projection of the scene, reconstructed from top-down RGB-D images.

$a_{\text{pick}}$ and $a_{\text{place}}$ are parameterized by the tuples $(u, v, \theta)$, indicating a top-down grasp or release at the pixel coordinates $(u, v)$, oriented at the angle $\theta$ around the gravitational axis. The coordinates $u$ and $v$, along with the orientation $\theta$, are discretized. Specifically, $\theta$ is defined within $\{2\pi i/N\}_{i=0}^{N-1}$, where N is the number of orientations allowed in the action space, which determines the orientation precision of actions.

### B. Fine-grained Orientation Estimation

Central to our approach is the fine-grained orientation estimation method. This method employs an SO(2)-equivariant network utilizing the SO(2) irreps together with a group discretization operation to generate discrete orientation distributions. This approach allows flexible control over orientation precision and ensures consistent computational demands across different levels of orientation granularity.

*1) Irrep Fields of SO(2):* While the SO(2) group possesses an infinite number of irreps, our approach selectively utilizes a finite subset to construct SO(2)-steerable kernels. Specifically, it employs a representation $\rho_{\text{irrep}}$ formed by the direct sum of base irreps $\rho_j$ up to a predefined cutoff frequency $j_c$, formalized as:

$$\rho_{\text{irrep}} = \oplus_{j=0}^{j_c} \rho_j. \quad (4)$$

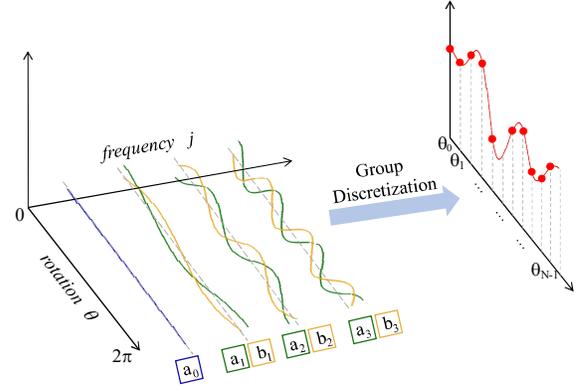

Fig. 2. Group discretization operation. Left: The input is a band-limited function in the Fourier domain, with coefficients $a_j$ and $b_j$ serving as parameters for the basis trigonometric functions. Right: The band-limited function is sampled at N orientations to produce a discrete orientation distribution, denoted by the red dots.

In this context, our network outputs an irrep field $\hat{f}: \mathbb{R}^2 \to \mathbb{R}^{d_{\text{irrep}}}$, assigning a coefficient vector $\hat{f}(x) \in \mathbb{R}^{d_{\text{irrep}}}$ to each spatial position $x \in \mathbb{R}^2$. These vectors encode band-limited functions in the Fourier domain, offering a continuous depiction of orientation probability distributions over a period of $2\pi$. An illustration of one such band-limited function is shown on the left side of Fig. 2. The dimensions of $\rho_{\text{irrep}}$ and $\hat{f}(x)$ are $d_{\text{irrep}} = 2j_c + 1$, aligning with the basis functions: $\cos(j\theta)$ and $\sin(j\theta)$ for frequencies $j \in [0, j_c]$. By employing the Inverse Fourier Transform (IFT), the band-limited function parameterized by each vector $\hat{f}(x)$ can be precisely sampled at any given angle $\theta \in [0, 2\pi)$:

$$\mathcal{F}^{-1}\left[\hat{f}(x)\right](\theta) = a_0 + \sum_{j=1}^{j_c}\left[a_j \cos(j\theta) + b_j \sin(j\theta)\right]. \quad (5)$$

Here, $a_0$, $a_j$ and $b_j$ are elements of $\hat{f}(x)$, associated with the specified basis functions, as depicted in Fig. 2.

*2) Group Discretization Operation:* To facilitate the training and optimization of the band-limited functions encoded in irrep fields, our method includes a group discretization operation, which transitions a continuous representation to a discrete one, as depicted in Fig. 2. It samples these functions at N distinct orientations distributed uniformly across SO(2), expressed as:

$$\text{Dis}_{C_N}^{\text{SO}(2)}(\hat{f}(x)) = \{\mathcal{F}^{-1}\left[\hat{f}(x)\right](2\pi i/N)\}_{i=0}^{N-1}, \quad (6)$$

where the IFT, defined in Equation 5, is applied to the coefficient vector $\hat{f}(x)$ at each designated orientation. The operation $\text{Dis}_{C_N}^{\text{SO}(2)}: \mathbb{R}^{d_{\text{irrep}}} \to \mathbb{R}^N$ essentially converts an irrep field of SO(2) to a regular field of $C_N$, transforming the coefficient vector $\hat{f}(x) \in \mathbb{R}^{d_{\text{irrep}}}$ at each position $x \in \mathbb{R}^2$ into a scalar vector $f(x) \in \mathbb{R}^N$ over N orientations. This discretization process is efficiently implemented via multiplication with an IFT matrix $Q \in \mathbb{R}^{N \times d_{\text{irrep}}}$, leading to negligible computational overhead. By adjusting the sampling number N, we achieve flexible control over orientation precision without requiring any modifications to the network structure. Note that the sampling number N must not be smaller than $d_{irrep}$ to prevent aliasing, adhering to the Nyquist-Shannon sampling theorem.

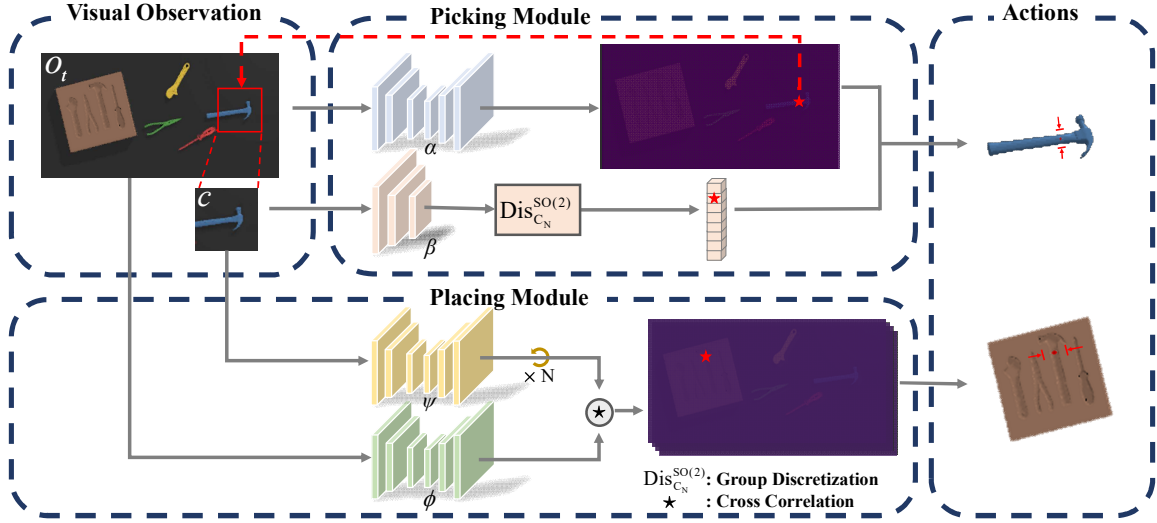

Fig. 3. Architecture of our kitting framework. This framework comprises a picking module and a placing module. The inputs for both modules include a scene observation $o_t$ and an image crop $c$ centered at the pick position. In the picking module, a network $\alpha$ predicts the pick position distribution, while a network $\beta$, followed by a group discretization operation, predicts the pick orientation distribution. In the placing module, two networks $\psi$ and $\phi$ encode $c$ and $o_t$ separately, with a cross-correlation operation determining the place action distribution. Optimal pick and place actions are selected by maximizing these distributions.

### C. Model Architecture

The architecture of our kitting framework is inspired by the Equivariant Transporter [17], comprising four equivariant convolutional networks: $\alpha$ and $\beta$ for the picking module, alongside $\phi$ and $\psi$ for the placing module, as depicted in Fig. 3. While achieving a level of equivariance similar to that of [17], our framework distinctively integrates the fine-grained orientation estimation method in the picking module and utilizes SO(2) irreps across all network designs.

*1) Picking Module:* The picking module operates in two stages to estimate the pick probability distribution $p(a_{pick} | o_t)$ by decomposing it into a positional component $p(u,v)$ and an angular component $p(\theta | (u,v))$. These components are evaluated sequentially using two distinct models: $f_p(o_t) \rightarrow p(u,v)$ and $f_\theta(o_t,(u,v)) \rightarrow p(\theta | (u,v))$.

The pick position model $f_p$ maps an observation $o_t \in \mathbb{R}^{4 \times W \times H}$ to a spatial probability map $p(u,v) \in \mathbb{R}^{1 \times W \times H}$ over potential picking locations directly. It incorporates an SO(2)-invariant network $\alpha$, which adopts a U-net structure enhanced with steerable convolutional layers. Both the input $o_t$ and the output $p(u,v)$ are defined as trivial fields to maintain invariance to rotational changes, while the intermediate layers leverage irrep fields to encode rotational symmetries. The optimal pick position $(u^*, v^*)$ is identified by maximizing $p(u,v)$ : $(u^*, v^*) = arg\max_{(u,v)}(p(u,v))$.

The pick angle model $f_\theta$ derives a probability vector $p(\theta | (u^*, v^*)) \in \mathbb{R}^N$ for N discrete orientations at the predetermined pick position. This model employs the fine-grained orientation estimation method by integrating an SO(2)-equivariant network $\beta$ with a group discretization operation $Dis_{C_N}^{SO(2)}$. The network $\beta$ takes an image crop $c_1 \in \mathbb{R}^{4 \times W_1 \times H_1}$ centered at $(u^*, v^*)$ as its input, producing an irrep vector $\beta(c_1) \in \mathbb{R}^{d_{irrep}}$ that models a continuous, unnormalized probability distribution over SO(2). Subsequently, this irrep vector is discretized at N orientations using $Dis_{C_N}^{SO(2)}$ and then passed to softmax to yield a discrete, normalized orientation distribution $p(\theta | (u^*, v^*))$. The optimal pick angle is identified by maximizing this distribution: $\theta^* = arg\max_\theta (p(\theta | (u^*, v^*)))$.

The use of a bilaterally symmetric gripper in planar picking introduces an additional symmetry, where the success rate of a grasp remains invariant when the gripper pose is rotated by $\pi$. To integrate this symmetry in our model, we utilize irreps of the quotient group $SO(2)/C_2$ in $f_\theta$, which groups orientations separated by $\pi$ into equivalent classes. Unlike the SO(2) group, this quotient group is realized using basis functions with a period of $\pi$. Accordingly, the group discretization operation is adapted to align with the quotient group, denoted as $Dis_{C_N/C_2}^{SO(2)/C_2}$. The adjustment halves the size of the output probability vector to $N/2$.

*2) Placing Module:* The placing module $f_{place}(o_t, c_2) \rightarrow p(a_{place} | o_t, a_{pick}^*)$ predicts place distribution by performing cross-correlation between the encoded observation $o_t$ and the rotated encodings of an image crop $c_2$ centered at $a_{pick}^*$. This process incorporates two SO(2)-invariant networks, $\phi$ and $\psi$, to encode the observation $o_t$ and the image crop $c_2$ separately. Both networks employ U-net structures similar to the pick position network $\alpha$. Afterwards, the encoded image crop $\psi(c_2)$ is subjected to N distinct rotations using a lifting operation $R_N$, resulting in a stack of N rotated encodings for $c_2$, expressed as:

$$R_N(\psi(c_2)) = \{T_{2\pi i/N}^{\rho_0}(\psi(c_2))\}_{i=0}^{N-1}. \quad (7)$$

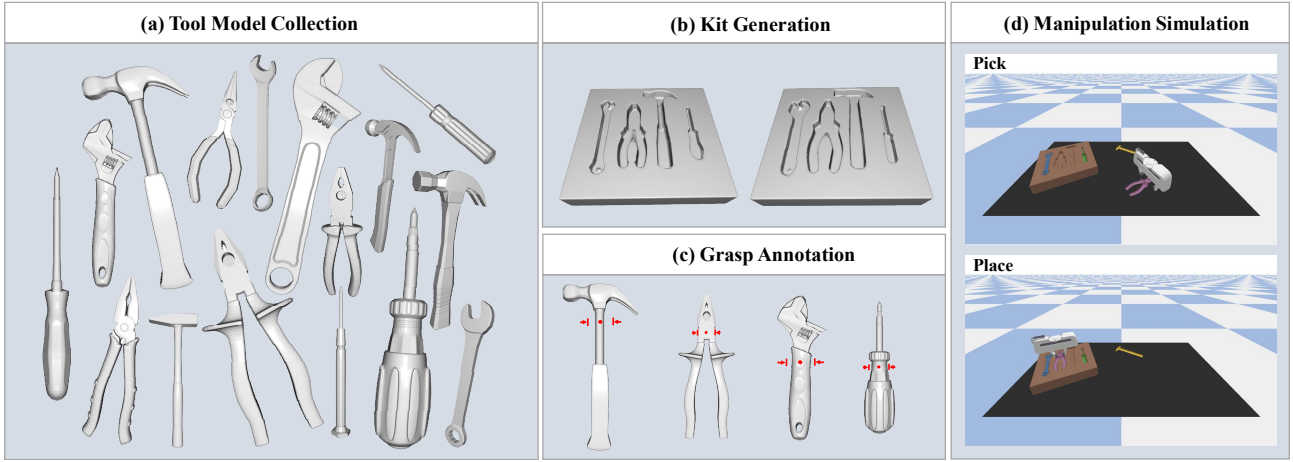

Fig. 4. Generation of the Hand-tool Kitting Dataset. The systematic pipeline for creating the dataset includes: (a) tool model collection, gathering four types of hand tools including hammers, pliers, wrenches and screwdrivers; (b) kit generation, creating synthetic kits with conformal cavities for each tool type; (c) grasp annotation, with red arrows indicating the annotated grasp pose for each tool type; (d) manipulation simulation, simulating a complete kitting process that includes both pick and place actions.

The place distribution is then determined by cross-correlating the encoded observation $\phi(o_t)$ with $R_N(\psi(c_2))$:

$$f_{\text{place}}(o_t, c_2) = R_N(\psi(c_2)) \star \phi(o_t). \quad (8)$$

The optimal place action $a_{\text{place}}^*$ is determined by maximizing $f_{\text{place}}$ across all pixel positions and orientation channels.

## IV. HAND-TOOL KITTING DATASET

### A. Design Principle

The HKD is meticulously developed to capture the challenges encountered in real-world hand-tool kitting tasks, with a particular focus on orientation sensitivity. This dataset is composed of a collection of common hand tools, each with a unique and irregular geometry that renders the kitting process highly susceptible to minor local misalignment. Characterized by their high aspect ratios, these tools underscore the necessity of precise orientation control, as even slight orientation deviations can result in considerable endpoint displacements. The design of the kits complements these tools with cavities that closely conform to their geometries with minimal clearance. Moreover, the HKD simulates physical interactions between tools and their cavities, realistically replicating scenarios where slight mismatches might allow a tool to slip into its position. By emphasizing geometric fidelity and physical realism, the HKD serves as a robust benchmark for evaluating algorithms on orientation-sensitive kitting tasks.

### B. Dataset Generation

The dataset is generated via a structured pipeline, as depicted in Fig. 4, comprising four steps: tool model collection, kit generation, grasp annotation and manipulation simulation.

*1) Tool Model Collection:* The HKD features four types of hand tools: hammers, pliers, wrenches and screwdrivers, each with fifteen distinct instances. These models are sourced from online 3D model communities, originally acquired either by scanning actual tools or crafting digital replicas. To assess the algorithm's ability to generalize to unseen objects, the instances in each category are divided into two groups: ten for training and five for testing.

*2) Kit Generation:* To address the shortage of kit models, we develop an automated kit generation process and create 100 different kits for both training and testing purposes. Our kit design follows a standardized template, specifying a fixed kit block size and predefined locations for cavities designed to accommodate four unique tools from separate categories. To ensure variety, we randomly select a tool from each category to form a diverse toolset. Additionally, the selected tools undergo random proportional scaling, with their lengths adjusted to fall in the [15cm, 20cm] range.

The creation of conformal cavities starts with each tool positioned flatly in a simulation environment, where a top-down orthographic depth image is captured to represent the top surface of the tool. This image is transformed into an occupancy grid by filling the volume beneath the top surface to match the actual height of the tool. A 3D morphological closing operation then smoothens the tool shape, and a uniform 2mm scaling is applied to ensure a slight clearance. The final step involves subtracting the adjusted tool geometry from the standardized kit block at its intended location, accurately crafting the conformal cavities.

*3) Grasp Annotation:* We manually label the grasping pose of each tool model using a customized annotation interface. To minimize grasping uncertainty, the annotation ensures the proximity to the center of mass and considers the flatness and parallelism of opposing surfaces. Fig. 4(c) illustrates the annotated grasping pose for each category. Given the predefined cavity locations in the kit block, the placement pose for each tool can be determined accordingly.

*4) Manipulation Simulation:* Our simulation of the hand-tool kitting task is implemented based on [17]. Each simulation round begins with four tools and their corresponding kit randomly positioned on a tabletop setup without collisions. The goal is to determine the pick and place poses for a Franka Emika gripper, enabling it to transfer all tools into their corresponding cavities in the kit block. We assume randomized manipulation sequence of the tools to introduce variability and challenge to the task. To facilitate data collection, an oracle agent is developed to perform expert demonstrations based on the ground-truth poses of both tools and kits, alongside the annotated pick and place poses.

*C. Evaluation Metric*

Our evaluation criteria are centered on the precise fit and complete filling of the cavities by the tools. To quantify this kitting outcome accurately, we adopt the average distance metric for symmetric objects (ADD-S) [24], a metric commonly used in pose estimation challenges. The ADD-S metric calculates the mean distance between the 3D model points $x_1$, transformed by its target pose $(\bar{\mathbf{R}}, \bar{\mathbf{t}})$ within the cavity, to their nearest points $x_2$ on the model transformed by the actual pose $(\mathbf{R}, \mathbf{t})$ after placement:

$$\text{ADD-S} = \frac{1}{m} \sum_{x_1 \in \mathcal{M}} \min_{x_2 \in \mathcal{M}} \left\| (\bar{\mathbf{R}} x_1 + \bar{\mathbf{t}}) - (\mathbf{R} x_2 + \mathbf{t}) \right\|. \quad (9)$$

Here, $m$ is the number of points in the 3D model's point set $\mathcal{M}$. The kitting of a tool is deemed successful if the average distance is below a predefined threshold of 3.5mm in our experiment setups.

## V. EXPERIMENTS

*A. Implementation Details*

Our framework is implemented using PyTorch and escnn [15] libraries. The types of groups and representations for our networks are outlined as follows:

*1) Networks $\alpha$, $\phi$ and $\psi$:* These networks are constructed using similar 21-layer U-net structures that leverage SO(2) irreps. Their representations are uniformly defined as a direct sum of irreps across frequencies $j \in [0,3]$, denoted as $\rho_{\text{irrep}} = \bigoplus_{j=0}^{3} \rho_j$.

*2) Network $\beta$:* This network adopts a 9-layer structure. The initial eight layers utilize SO(2) irreps, while the final layer transitions to an irrep of the quotient group $SO(2)/C_2$. Despite the group difference, their representations are uniformly defined within the same frequency domain of their respective irreps: $\rho_{\text{irrep}} = \bigoplus_{j=0}^{6} \rho_j$.

*B. Tasks*

*1) Hand-Tool Kitting Tasks:* We utilize the proposed HKD as the benchmark for evaluating the performance of our kitting framework. To assess its performance on both seen and unseen tools, two distinct task setups are introduced:

- **kitting-seen-toolset:** a predetermined set of hand tools and their corresponding kit are consistently used throughout training and testing.
- **kitting-unseen-toolsets:** a pair of toolset and kit is randomly selected from separate pools for each training and testing round.

Performance evaluations are conducted across a range of orientation numbers, N = 36, 72, 120 and 180, representing different orientation precisions. Each model is consistently trained and evaluated with the same N value.

*2) Manipulation Tasks with Lower Orientation Sensitivity:* Five modified tasks from Raven-10 [8], [17] are utilized to evaluate our framework's ability to handle a variety of tasks beyond kitting. These tasks, tailored for a parallel-jaw gripper, include block-insertion, align-box-corner, place-red-in-green, stack-block-pyramid and palletizing-boxes. We leverage them to assess our algorithm's adaptability to tasks with lower orientation sensitivity by setting N = 36 initially. Additionally, we extend the evaluation to N = 180 to explore the impact of increased orientation granularity on these tasks.

*C. Training and Evaluation*

*1) Training:* A dataset of *n* expert demonstrations is generated for each task, where *n* varies based on the task's orientation sensitivity. For hand-tool kitting tasks, *n* is set to 10 and 100. For the modified Raven-10 tasks, *n* is set to 1, 10 and 100. Each demonstration comprises a sequence of one or more observation-action tuples $(o_t, \bar{a}_{\text{pick}}, \bar{a}_{\text{place}})$. The expert actions, $\bar{a}_{\text{pick}}$ and $\bar{a}_{\text{place}}$, are encoded into one-hot pixel maps as the ground-truth labels. We employ cross-entropy loss to train our models, utilizing the Adam optimizer for 10k iterations with a batch size of 1 and a learning rate of $10^{-4}$. Evaluations are conducted every 2k steps on 100 unseen test runs. All the training and evaluation are performed on a computing server with an AMD Ryzen Threadripper PRO 3995WX CPU and an NVIDIA RTX 3090 Ti GPU.

*2) Evaluation Metrics:* Our evaluation assesses both the success rate and computational efficiency of each model. We measure success rate on a scale from 0 (failure) to 100 (success), awarding partial scores for tasks that involve multiple actions. The reported results reflect the highest validation performance attained during training, with an average taken over 100 unseen test trials for each task. For computational efficiency, our evaluation is made based on the total number of parameters and the inference time per test run.

*3) Baselines:* Our evaluation involves comparison against two primary baselines. The first, Equivariant Transporter [17], is a variation of our method, utilizing $C_N$ regular representations in its networks. It employs a specific setup by using $C_6$ in $f_p$, $\phi$ and $\psi$, and $C_N/C_2$ in $f_\theta$. The second is adapted from Transporter Net [8], integrating three 43-layer ResNets built on conventional convolutional layers. This adaptation is designed for parallel jaw grippers by directly rotating input scene images, producing a stack of oriented images as the input of its $f_{\text{pick}}$ network.

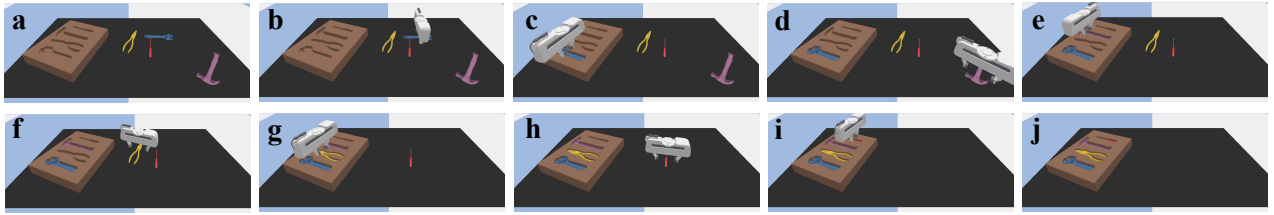

Fig. 5. Manipulation sequence in an evaluation round of the kitting-unseen-toolsets task. (a) shows the starting state of the kitting task; (b)-(i) illustrate the sequence of manipulations to kit the hand tools; and (j) depicts the completed state of the kitting task.

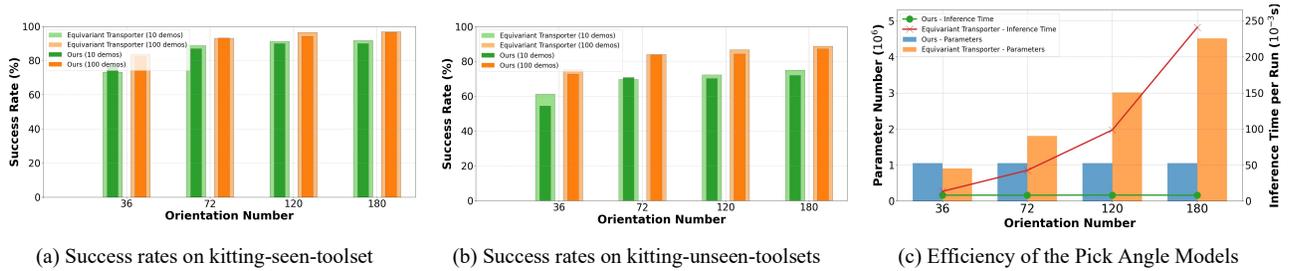

(a) Success rates on kitting-seen-toolset
(b) Success rates on kitting-unseen-toolsets
(c) Efficiency of the Pick Angle Models

Fig. 6. Comparative performance on hand-tool kitting tasks. This figure offers a detailed comparison of our method against the Equivariant Transporter across various orientations and demonstrations. (a) and (b) show the success rates for tasks with seen and unseen toolsets, respectively. Green bars denote models trained with 10 demonstrations, and orange bars are those trained with 100 demonstrations. Dark, slim bars represent our methods, while the overlaid lighter, wide bars correspond to the Equivariant Transporter, under identical configurations. (c) contrasts the inference time per test run and the number of parameters in their pick angle models, with lines indicating inference time and bars showing parameter numbers.

*D. Results: Hand-Tool Kitting Tasks*

Fig. 5 illustrates a successful sequence of kitting unseen tools into their designated cavities using our proposed method, configured with 180 rotations and 100 demonstrations. In this sequence, the tools are accurately kitted in a randomized order, as directed by our model, and the kitting strategies are dynamically adapted to different tools. A detailed quantitative analysis is provided below.

*1) Success Rate:* Fig. 6(a) and 6(b) present the success rates of our method versus the Equivariant Transporter, measured across varying configurations. The results highlight the effectiveness of both strategies in solving kitting tasks, particularly with adequate orientation precision and demonstrations. Our approach matches the baseline's success rates closely, with a minor average decline of 0.63% on a seen toolkit and 2.16% on unseen toolkits. Specifically, our approach achieves a success rate of 96.5% for a seen toolkit and 87.25% for unseen toolkits with a setup of 180 rotations and 100 demonstrations.

An improvement in success rates is observed with larger orientation numbers for both methods, underlining the importance of precise orientation control in kitting tasks. Meanwhile, this finding also affirms the HKD as a reliable benchmark for evaluating solutions for orientation-sensitive kitting tasks. Besides, the notable performance disparity between tasks involving seen and unseen tools reflects the inherent challenge of generalizing to new objects.

*2) Computational Efficiency of the Picking Module:* Our comparative study with the Equivariant Transporter reveals parallel efficiency levels in their placing modules, attributed to their similar designs. Thus, we shift our focus toward the computational efficiency of the picking modules.

As the pick position models $f_p$ of both methods are designed to be invariant to orientation changes, they exhibit consistent efficiency throughout the experiments. Each model maintains a parameter count of 4.54 million and an inference time of 0.06 seconds. However, there is a notable difference in the computational efficiency of their pick angle models $f_\theta$, as shown in Fig. 6(c). Our angle model maintains consistent parameter counts and inference times across various orientations. In contrast, the pick angle model of the Equivariant Transporter scales linearly in parameters with the orientation numbers, and its inference time also grows with larger orientation counts. Thus, the uniform computational efficiency of our pick position and angle models ensures the overall efficiency of our picking module remains unaffected by variations in orientation precision.

*3) Discussion:* The divergence in efficiency between pick angle models arises from their distinct strategies for managing orientation granularity. Our model utilizes a consistent set of SO(2) irreps regardless of orientation number, with granularity adjusted by a group discretization operation. Conversely, the Equivariant Transporter explicitly models probability distributions over N orientations using $C_N$ regular representations. While slightly improving success rates, this strategy requires an expanded set of irreps in the underlying structure of the network as orientation granularity increases, thus elevating computational demands.

Moreover, the orientation precision for place actions is governed by the rotation count of encoded image crops in the placing module. Although increasing rotation numbers leads to more computations due to the cross-correlation operation, the image encoding networks remain unaffected. Consequently, our entire kitting framework maintains consistent model parameter numbers irrespective of orientation granularity. This consistency underscores our framework's ability to achieve diverse orientation precision without modifying the underlying network structures, showcasing its versatility and efficiency.

TABLE I. Success Rates on Modified Raven-10 Tasks With A Parallel-Jaw Gripper

| Method | block-insertion | | | place-red-in-green | | | palletizing-boxes | | | align-box-corner | | | stack-block-pyramid | | |
|---|---|---|---|---|---|---|---|---|---|---|---|---|---|---|---|
| | 1 | 10 | 100 | 1 | 10 | 100 | 1 | 10 | 100 | 1 | 10 | 100 | 1 | 10 | 100 |
| Ours-36 | **100** | **100** | **100** | 96.0 | **100** | **100** | **97.8** | 99.9 | **100** | 58.0 | **100** | **100** | 51.8 | 84.5 | 97.5 |
| Equivariant Transporter | **100** | **100** | **100** | 95.6 | **100** | **100** | 96.1 | **100** | **100** | **64.0** | 99.0 | **100** | **62.1** | 85.6 | 98.3 |
| Transporter Network | 98.0 | **100** | **100** | 82.3 | 94.8 | **100** | 84.2 | 99.6 | **100** | 45.0 | 85.0 | 99.0 | 16.6 | 63.3 | 75.0 |
| Ours-180 | 95.0 | **100** | **100** | **98.2** | **100** | **100** | 96.2 | **100** | **100** | 42.0 | 95.0 | **100** | 56.0 | **86.8** | **99.0** |

*E. Results: Modified Raven-10 Tasks*

Table I presents the success rates of our approach on the modified Raven-10 tasks, detailing performance at two levels of orientation granularities: N = 36 and 180.

*1) Task Generalization:* Starting with N = 36 as the baselines, our method attains state-of-the-art results on nearly all tasks, except for a slight underperformance in the stack-block-pyramid task. The results highlight its adaptability to diverse manipulation tasks with varying orientation sensitivities. Moreover, our approach exhibits high sample efficiency, performing well on most tasks with a small number of demonstrations.

*2) Effects of High Orientation Precision:* After elevating the orientation number to N = 180, our method maintains high success rates. This is particularly evident in its performance on the challenging stack-block-pyramid task, where enhanced orientation precision allows our approach to outperform all others.

## VI. Conclusion

This study presents an in-depth exploration of orientation-sensitive kitting tasks, which leads to the development of a novel kitting framework and the Hand-tool Kitting Dataset. By integrating a fine-grained orientation estimation method into our picking module, we have significantly improved both orientation precision and computational efficiency in kitting operations. The HKD emerges as a reliable benchmark for evaluating solutions against orientation-sensitive kitting tasks, mirroring the complexities of real-world kitting scenarios. Our extensive experiments validate the remarkable efficiency and adaptability of our framework. Future efforts will extend to conducting physical experiments to validate our approach in real-world hand-tool kitting scenarios. Additionally, we aim to enhance the success rate of our system on the HKD further, while ensuring sustained operational efficiency.